# Challenges and Opportunities of Speech Recognition for Bengali Language

**M. F. Mridha**
Department of Computer Science & Engineering
Bangladesh University of Business & Technology
Dhaka, Bangladesh
firoz@bubt.edu.bd

**Abu Quwsar Ohi**
Department of Computer Science & Engineering
Bangladesh University of Business & Technology
Dhaka, Bangladesh
quwsarohi@bubt.edu.bd

**Md. Abdul Hamid**
Department of Information Technology
Faculty of Computing & Information Technology
King Abdulaziz University
Jeddah-21589, Kingdom of Saudi Arabia
mabdulhamid1@kau.edu.sa

**Muhammad Mostafa Monowar**
Department of Information Technology
Faculty of Computing & Information Technology
King Abdulaziz University
Jeddah-21589, Kingdom of Saudi Arabia
mmonowar@kau.edu.sa

## Abstract

Speech recognition is a fascinating process that offers the opportunity to interact and command the machine in the field of human-computer interactions. Speech recognition is a language-dependent system constructed directly based on the linguistic and textual properties of any language. Automatic Speech Recognition (ASR) systems are currently being used to translate speech to text flawlessly. Although ASR systems are being strongly executed in international languages, ASR systems' implementation in the Bengali language has not reached an acceptable state. In this research work, we sedulously disclose the current status of the Bengali ASR system's research endeavors. In what follows, we acquaint the challenges that are mostly encountered while constructing a Bengali ASR system. We split the challenges into language-dependent and language-independent challenges and guide how the particular complications may be overhauled. Following a rigorous investigation and highlighting the challenges, we conclude that Bengali ASR systems require specific construction of ASR architectures based on the Bengali language's grammatical and phonetic structure.

## 1 Introduction

Undoubtedly, speech is the most fascinating and competent form of interaction among one another. Moreover, it is additionally conceivable to utilize speech as an outstanding medium to interact with machines. Wherefore, speech recognition investigation has advanced from research center exhibitions to genuine applications. Hence, speech recognition systems are frequently observed and accepted in daily use applications [1]. This daily usage and dependence on ASR systems require the architecture to be accurate to its best. A user may feel interrupted if the ASR-based search system outputs scrambled or wrong words while he/she is using a voice search feature on the web or calling the wrong person while using ASR-based automated calling functions. Hence, implementing an accurate ASR system requires an in-depth analysis of the speech-to-text translation systems, including grammar and word-level knowledge.

Table 1: The table exhibits some of the surveys conducted in the domain of ASR systems. Most of the surveys cover a specific event of the ASR systems.

| **Study** | **Reviewed Feature Extraction Strategies** | **Reviewed Deep Learning Strategies** | **Discussed Existing ASR Methods** | **Reviewed Datasets** | **Discussed Grammatical Variation** | **Core Contribution** |
|---|---|---|---|---|---|---|
| [2] | ✗ | ✗ | ✓ | ✗ | ✗ | Comparing HMM and ANN architectures. Pointing towards the performance improvement of hybrid architectures. |
| [3] | ✓ | ✗ | ✓ | ✗ | ✗ | Discussion in speech recognition based on speech variations, such as emotion, phycology, speech rate, accent, etc. |
| [4] | ✗ | ✗ | ✓ | ✗ | ✗ | Through discussion of speech recognition in adverse conditions. |
| [5] | ✓ | ✗ | ✓ | ✓ | ✗ | Focused on under-resourced languages. Discussed extinction, challenges, and resources of such languages. |
| [6] | ✓ | ✗ | ✗ | ✗ | ✗ | Addressed the advantages and disadvantages of classic ASR techniques. |
| [7] | ✓ | ✓ | ✓ | ✓ | ✗ | Brief in Turkish speech recognition. |
| **Ours** | ✓ | ✓ | ✓ | ✓ | ✓ | Brief in Bengali speech recognition along with architectural strategies concerning grammatical properties. |

Language dependency is one of the greatest obstructions of a speech recognition system. Thus a speech recognition system has to target a specific language base. Due to language dependency, a system that better recognizes English speech may not correctly recognize other linguistic speech. Moreover, language dependency is solely due to the grammatical properties of specific languages. A similar condition also applies to the Bengali language, which has wider structural and grammatical variations than the English language. However, language dependency has not been frequently investigated by researchers. Apart from discussing existing literature of ASR systems and datasets, most research has been conducted in algorithm selections, speech variation challenges, architectural investigation, etc. Table 1 represents a comparison of some of the notable and recent analyses conducted in the ASR literature. Consequently, in this research endeavor, we deeply investigate the grammatical aspects of speech recognition, along with the challenges of algorithms w.r.t. grammar and phones.

In this paper, we ground our discussion on the challenges and opportunities that a Bengali ASR system pose. The core contribution of the paper includes:

- We conduct a comprehensive investigation of most of the works undertaken in the Bengali ASR system, including speech corpora and architecture. To the best of our knowledge, no comprehensive survey has been made discussing the grammatical and architectural relation of ASR systems.
- We point out various challenges encountered while implementing Bengali ASR systems. Moreover, we provide an anatomy of the challenges and discuss linguistic and grammatical differences between English and Bengali language.
- Finally, we provide future directions that should be recollected while building architectures. Further, we provide an optimal structure that may resolve the issues of Bengali ASR systems.

The rest of the paper is segmented as follows. Section 2 acquaints the generic architectures that are investigated in the ASR domain. Section 3 introduces the attempts which are executed in the track of the Bengali ASR system. Section 4 contains a detailed investigation of the challenges Bengali ASR system poses. Section 5 summarizes the overall challenges introduced in the paper and proposes an optimal architecture to solve the challenges. Finally, Section 6 concludes the paper.

## 2 Attempts in ASR System

The first speech recognition system was introduced in 1920, which was the first machine to recognize speech [8]. Later, the journey of speech recognition technology continued to be improved by the independent works of researchers all around the globe. The researchers interested in speech recognition systems introduced and adopted many state-of-the-art techniques that have improved and are still improving the precision of speech recognition systems. Pattern matching approaches such as brute-force techniques, phonetic segmentation, and hybrid systems were first introduced in speech recognition systems. However, the vast improvement is often concerned after Hidden Markov Models (HMM) adaption, which appeared in late 1970. HMM has become popular in ASR systems due to more immeasurable pattern analysis expertise over large vocabularies [9, 10] and being feasible to practice [11].

Lately, after the improvement of Artificial Neural Network (ANN) architectures, neural network-based speech recognition systems are also proved and considered to be better. Popular architectures of Deep Neural Networks (DNN) such as Convolutional Neural Networks (CNN) [12], Residual Networks [13] are being implemented in ASR systems, and they are proving to be effective. DNN based architectures are also proven to be more effective than any other architectures implemented in the Bengali ASR system [14]. Popular feature extraction techniques like Principal Component Analysis (PCA) [15], Linear Discriminate Analysis (LDA) [16], Independent Component Analysis (ICA) [17], Wavelet Analysis [18] has been implemented to extract speech features from acoustic waveforms. Among the aforementioned feature extraction strategies, PCA is used to exact structure from input data. However, the drawback of PCA is that it can only recognize the linearity of data. In contrast to PCA, a deep learning-based strategy, named AutoEncoder, can acknowledge data's non-linearity. Hence, currently, AutoEncoders are being implemented to embed the non-linearity of data. In the case of LDA, a probabilistic LDA (PLDA) is used mostly to recognize features from speech embeddings. Both LDA and PDA are intensely studied in speaker recognition tasks [19, 20]. Specialized feature extraction systems like Mel-frequency Cepstrum Coefficient (MFCC) [21, 22], Cepstral Mean Subtraction [23], RASTA filtering [24, 25] is also observed to be used to extract features from waveforms. MFCC has been deeply investigated in the domain of speech and speaker recognition. Currently, MFCC is fused with various CNN architectures and mostly generating better accuracy in speech recognition frameworks. The reason for achieving better accuracy lies behind the mel-scales of the MFCC. A low-scale MFCC excludes unwanted features and greatly focuses on the phones of speech [26].

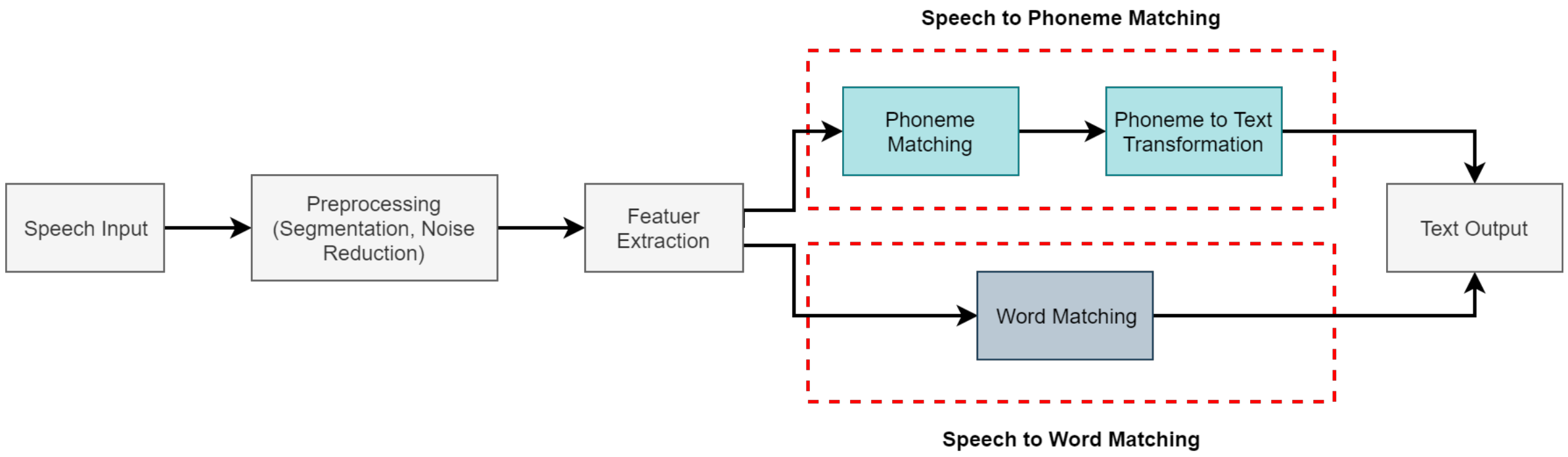

Figure 1: The figure illustrates the common architecture of automated speech recognition systems. The red dashed boxes frames the two basic pattern matching or classification schemes (phoneme and text matching) that are frequently practiced in speech recognition architectures.

An ASR system has two principal processing architectures, which are observed mostly to exist, a) Feature Extraction and b) Pattern Matching. Feature extraction is the process of extracting speech parameters having acoustic correlation from an acoustic waveform [27], and pattern matching is the process of matching the extracted speech features with the correct output from the template database [28]. The pattern matching can be either speech to phoneme matching [29], or speech to word matching [30], although we define a hybrid method that can perform both. Generally, the term hybrid is mostly used to identify such ASR architectures that combine HMM and Multi-Layer Perceptron (MLP) method [31, 32]. However, in this paper, we define the term hybrid on the basis of the combination of speech to text and speech to phoneme scheme. The proper combination and tweaks applied to the two principal architectures (feature extraction and pattern matching) may significantly increase the performance of the system. However, some attachments such as word segmentation (segmenting speech frames from continuous speech), noise reduction, and phoneme to word transformation can be observed in ASR systems to enhance the usage and robustness of ASR systems. Figure 1 demonstrates the overall course of processes which are performed in an ASR system. Furthermore, Figure 2 demonstrates the overall course of processes that are performed in a hybrid ASR system.

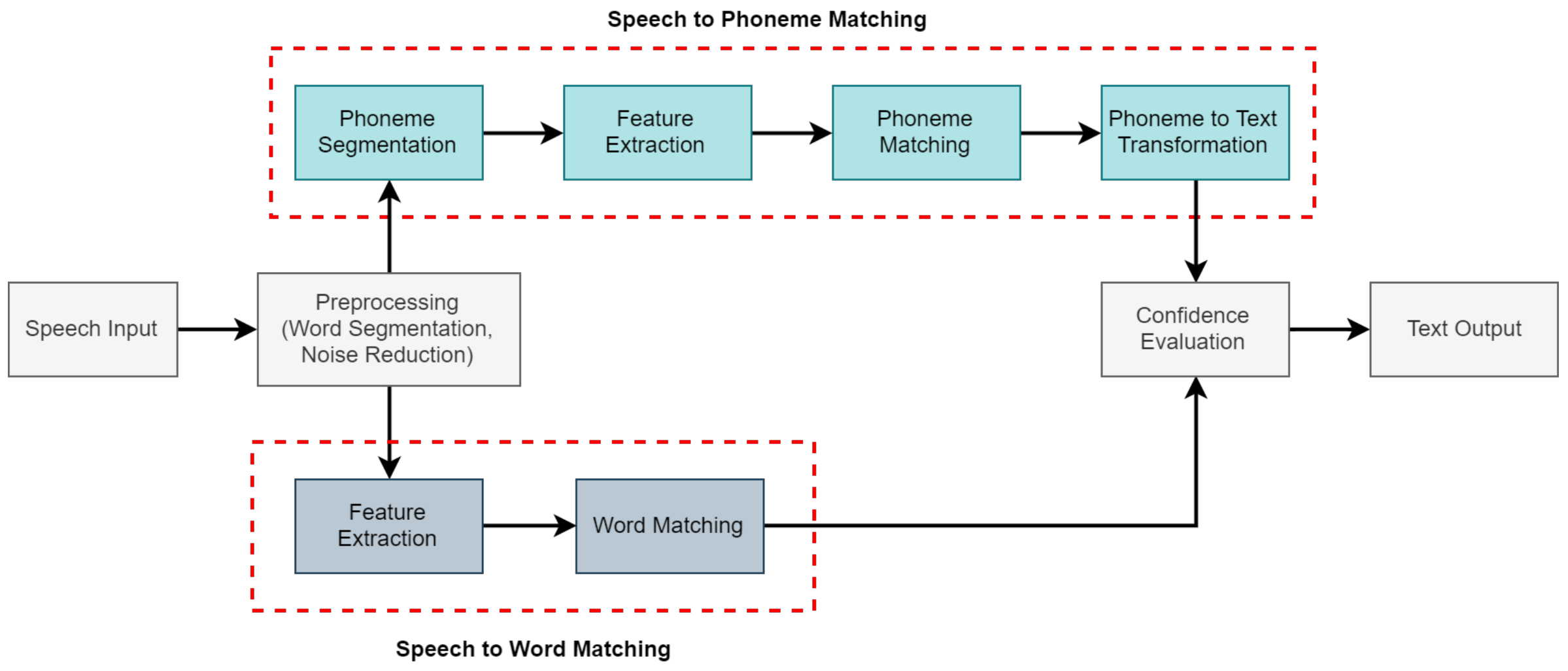

Figure 2: The figure illustrates the standard architecture of hybrid automated speech recognition systems. The red-dashed boxes frame the two essential pattern matching, or classification schemes (phoneme and text matching) frequently practiced in speech recognition architectures. The final text output is obtained based on the confidence evaluation of the word matching scheme.

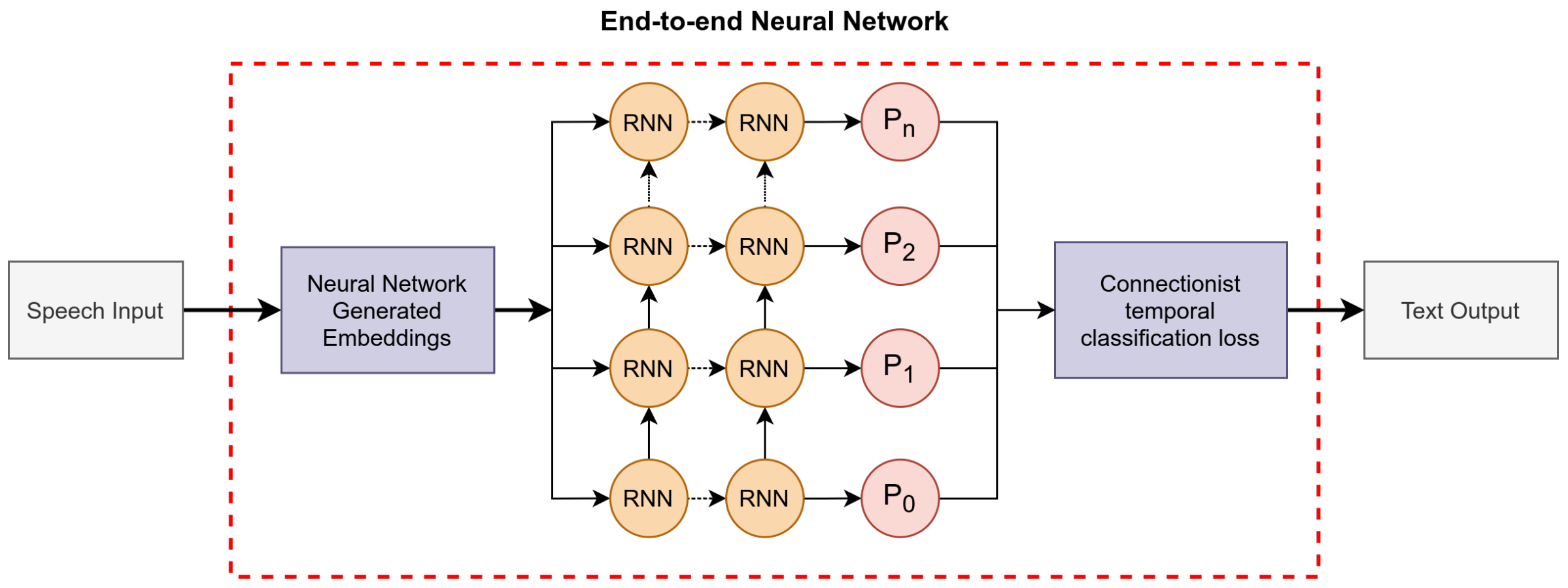

Figure 3: The figure illustrates a general end-to-end structure for speech recognition. The neural network generates embeddings from input features and further passes them to a stack of recurrent layers. The recurrent layers find patterns based on previous and current input features and generate a final output. The CTC loss is used to train the network via backpropagation.

Apart from the general strategies of speech recognition, the current improvement of or recurrent neural networks (RNN) has led the speech recognition system to a new strategy named end-to-end ASR [33]. A single RNN based architecture performs feature extraction and speech to pattern matching simultaneously in an end-to-end method. The advantage of the end-to-end strategy is that the whole network is always trained using a single loss function. Connectionist temporal classification (CTC) loss is broadly implemented as a loss function in an end-to-end framework. However, the limitation of these methods is that they require a considerably large amount of data to work precisely [34]. Moreover, it also requires a considerable amount of time to attain optimal features from the input stream. Figure 3 visualizes the basic structure of an end-to-end framework.

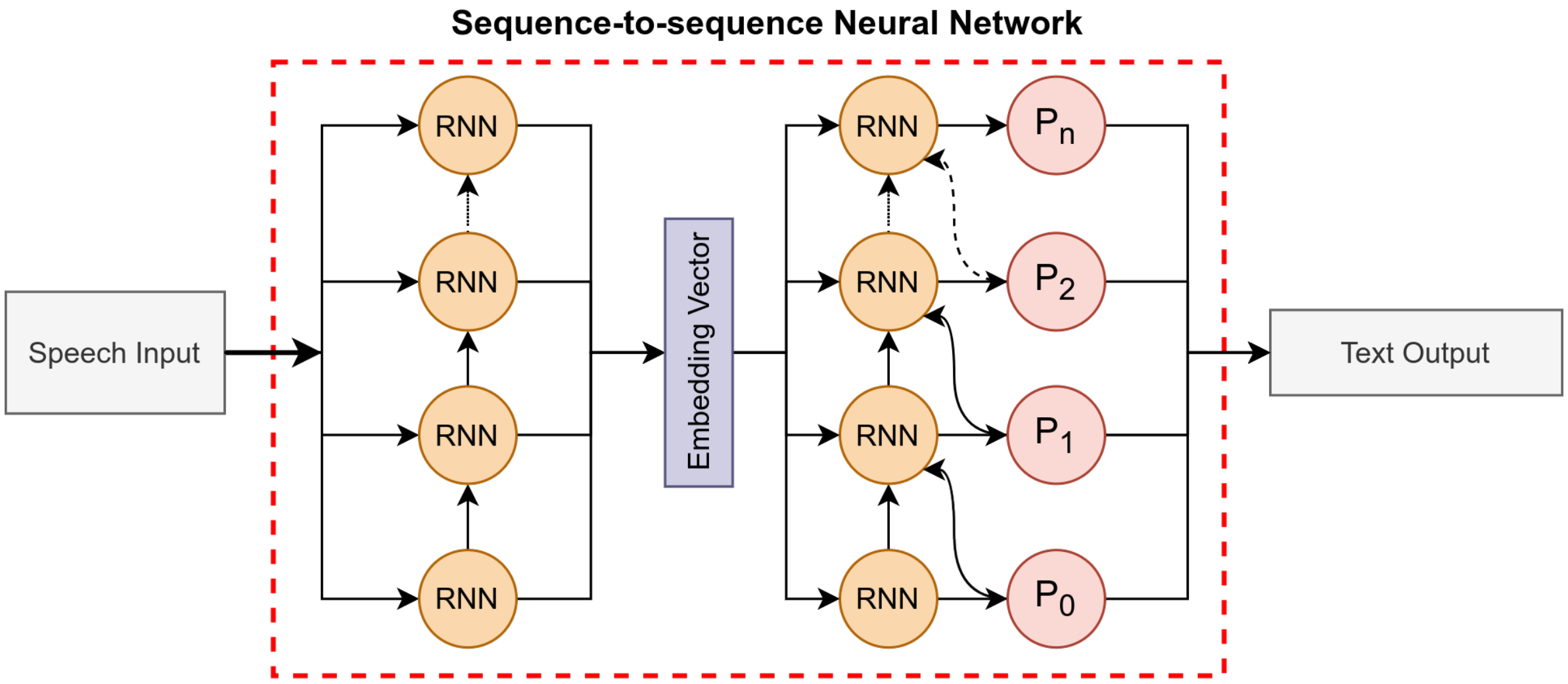

Figure 4: The figure visualizes a general sequence to sequence model for speech recognition. The seq2seq network contains an encoder consisting of a stack of RNN that produces embedding vectors. The decoder comprising RNN receives the embedding vectors and produces final results. However, the RNN has access to the previous prediction. Therefore, the subsequent predictions have the possibility of being more accurate.

Some modifications of the end-to-end architectures have proven to be remarkably suitable for continuous speech and text processing. Amongst them, sequence-to-sequence (seq2seq) and attention-based models are well-considered. Seq2seq models contain an encoder and a decoder, both having a stack of RNN layers. The encoder generates meaningful embeddings from the input and encourages the decoder towards correct predictions. Figure 4 illustrates a common scenario of seq2seq framework. On the contrary, attention-based architectures [35] perform similarly as a seq2seq model [36]. Specifically, the attention mechanism is attached to a seq2seq model that extends the knowledge of previous inputs and outputs, resulting in a superior assumption of the network.

RNN has significantly been investigated in end-to-end architectures. As a result, two sophisticated strategies have been introduced, Long-short term memory (LSTM) [37] and Gated recurrent units (GRU) [38]. General RNN based architectures are prone to vanishing gradient problems, whereas LSTM and GRU networks evade such issues. LSTM and GRU both contain a memory of the previous states and have been preferred over general RNNs. GRU network requires fewer parameters in comparison to LSTM. However, LSTM has been proven to perform better in language modeling for speech recognition [39]. Recurrent architectures are still a region of interest to ASR researchers due to recognizing complex sequences from speech inputs.

## 3 Attempts in Bengali Speech Recognition

### 3.1 Attempts in Generating Bengali Speech Corpora

Efforts have been made in the Bengali speech recognition system, although there is still plenty to explore. Most of the works carried out in Bengali ASR systems are dispersed due to the absence of dataset availability. The scarcity of Bengali speech data caused the individual researchers to create their speech corpora, which has also not been made public. Therefore, most works were incomparable to each other, and it was impossible to prove the authenticity and quality of the corpora as well as research works. Currently, to the best of our knowledge, nine corpora are available for Bengali ASR systems. One is a real number speech corpus, one voice command corpus, and the others are full Bengali speech corpora. A complete analysis of the speech corpora is presented in Table 2.

The scarcity of Bengali speech datasets can only be resolved by producing massive, publicly available quality datasets. A quality speech recognition dataset has various usability domains, including speech to text processing, text to speech processing, speaker recognition, far-field speech recognition, etc. [54]. While creating a Bengali speech dataset, the following cases should be considered:

- Currently, quality speech datasets target specific environments: clean environment, telephony environment, broadcast (tv/radio) environment, meeting environment, far-field environment, in-the-wild environment. The most challenging environments are telephony, far-field, and in-the-wild environments. Most of the state-of-the-art speech recognition systems target these types of datasets.
- A Bengali speech recognition dataset should contain an accurate transcript of the speech. Also, it may contain speaker information (gender/emotion), environmental information as well.
- Diverse features in the speech dataset are required to make the dataset more challenging and practical. Diversity can be achieved in various constraints: input device, dialect, age, environment, noise constraint, speech disability, etc.
- Most famous datasets maintain clean and noisy datasets separate [55]. Separating clean and noisy datasets help researchers to implement Bengali speech recognition prototypes based on a particular scenario.
- A Bengali speech dataset should target Bengali-specific features, such as collecting speech from different dialects, collecting speech for critical and similar words, and especially handling the letter utterance similarity.

- A Bengali speech dataset must cover a large volume of word database and adequately present the statistics of the dataset variations.

Table 2: The table contains an insight into the currently available corpora suitable for Bengali speech recognition. The column 'Type' defines the category of the corpora. The 'Source' column explains the source from which the data was collected. 'Speech Length' column refers to an approximate length of the speech corpora in hours. 'Unique Utterances' column generates an approximate unique number of materials available in the corpora. The 'Repository Reference' column contains the reference link where the dataset can be found. The "Availability" column perceives whether the datasets are publicly or privately available.

| Collector | Type | Source | Speech Length | Unique Utterances | Repository Reference | Availability |
|---|---|---|---|---|---|---|
| [40] | Speech Phoneme | Voluntary Contribution | - | 47 phonemes | [41] | Public |
| [42] | Spoken Number Corpus | Voluntary Contribution | 3.8 hours | 115 numbers | [43] | Public |
| [44] | General Speech Corpus | Voluntary Contribution | 24 hours | - | [45] | Public |
| [46] | General Speech Corpus | Voluntary Contribution | 25 hours | 11,000 words | [47] | Public |
| [48] | Speech Command Corpus | Voluntary Contribution | - | 30 words | - | Private |
| [49] | General Speech Corpus | Voluntary Contribution | 26 hours | 19,640 words | - | Private |
| [50] | General Speech Corpus | Telephone Conversation | 215 hours | - | [51] | Private |
| [52] | General Speech Corpus | Crowd Sourced | 229 hours | 200,000 words | [52] | Public |
| [53] | General Speech Corpus | TV News, Audiobooks | 960 hours | 1,600,000 words | - | Private |

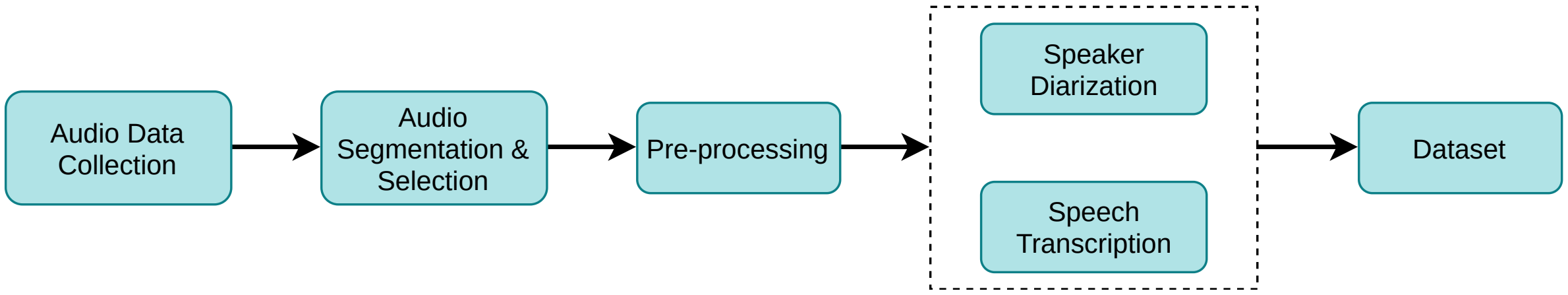

Figure 5: The image illustrates the general steps of preparing a Bengali speech dataset.

Creating a Bengali speech dataset specifically for deep learning architectures is challenging, as training the current deep learning strategies requires vast data. Figure 5 illustrates the general stages of the data collection procedure. The collection of speech datasets may include crowd-sourcing or an especially selected population. However, big datasets are often crowd-sourced. Further, a Bengali speech dataset may require additional statistical analysis to balance variation in numerous domains. It may require pruning and selection process as well. Moreover, the speech dataset requires some pre-processing, such as noise cancellation (optional), sound normalization, reducing silent intervals, and so on. Moreover, a manual or semi-automated process is to be conducted to generate speaker diarization and speech transcription. Finally, after validating the overall process, a quality Bengali speech dataset can be produced.

### 3.2 Attempts in Designing Bengali ASR Systems

Research works in the scope of the Bengali ASR system began in late 2000. Recognition of Bengali spoken letters [56] was introduced in an earlier stage. The pioneer works in the sector mainly were based on self-made short datasets and used statistical approaches [57, 58, 59, 60, 61]. The first work using Neural Networks, which was witnessed in 2009 [62]. The authors first pre-processed the input speech using pre-emphasis and hamming window. Then a 12 dimensional Linear Predictive Coding (LPC) is used to produce speech features. Finally, the speech features are fed to Artificial Neural Network to identify speech. However, the research work was conducted using a limited dataset of four persons, and no evidence of performance measurement was included.

In the following year, a continuous Bengali speech-to-text system was introduced. The work was carried out using CMUSphinx [63] (a speech recognition system), and a custom dataset was used to train the speech recognition system [64]. The system was designed using a phoneme pattern matching scheme and performed a phoneme to text translation system using tri-gram. CMUSphinx implements a three-state (tri-gram) statistical HMM and it uses GMM for probability distribution function. The approach generated 13% word error rate (WER) on 100 sentences.

In the same year, a speech segmenting method was also introduced that could segment Bengali speech from a continuous waveform [65]. The authors implemented mean windows to segment each of the words from continuous speech. Then, each segmented word was further referred among three clusters, belonging to mono, di, and tri syllable, based on the gaps in each segmented word. With six speakers of a 120 sentence dataset, the authors gained 98.48% accuracy.

The course of study continued, and in 2012, two new methods were introduced, among which, the first method was implemented using Microsoft Speech Application Programming Interface (SAPI) [66]. Due to the dependency on SAPI, the research work had a limitation. The architecture had to translate SAPI outputted English words to Bengali, and it was done through a direct English-to-Bengali word matching scheme. Therefore, the method fails to construct a Bengali word if it was not present in the English-to-Bengali word dataset. The second research work claimed that speech recognition systems might have an adverse effect depending on the gender of the speaker [67]. The research work was conducted using a self-made speech corpus and introduced an MFCC and HMM-based ASR architecture. It finally concluded that the ASR system performs better if both male and female speeches are present in the training samples.

A continuous speech to word pattern matching method was introduced in 2013, which was implemented using MFCC, Linear Predictive Coding (LPC), Gaussian Mixture Models, and Dynamic Time Wrapping (DTW) [68]. The authors implemented four different models each with different feature extraction and pattern matching scheme: a) MFCC + DTW, b) LPC + DTW, c) MFCC + GMM, d) MFCC + DTW. Among the four different setups, MFCC+GMM performed best by achieving 84% accuracy. However, the research work was conducted on a self-made dataset, and no comparison is performed. Further, due to the speech to word matching policy, the method may fail to recognize unknown meaningful and meaningless words.

The usage of DNN was first observed in 2017 that was a phoneme classification architecture [69]. The authors compared DNN and HMM architectures and proved DNN to be the most accurate. The DNN implementation contained stacked denoising autoencoders that took MFCC as input, which is pre-trained. Further, after pre-training the autoencoders, a multi-layer perceptron of three layers has been used to predict the phoneme probabilities. The baseline achieved 82.5% phoneme classification accuracy in a self-made dataset, which is unavailable. The authors also introduced a similar approach for classifying the place of speech sound articulation using DNN and AutoEncoders [70].

Table 3: The table represents the research efforts that are conducted in Bengali speech recognition. The 'domain' column explains the target of the research. The 'matching scheme' represents whether the pattern matching is performed using speech to word matching, or speech to phoneme matching. The 'features' column defines the feature extraction method of the proposed architecture. The column 'recognition method' explains the type of architecture that is used in the research work. The 'dataset' and 'accuracy' column represents the dataset that is used to train the model, and the test accuracy of the model, respectively.

| **Author** | **Domain** | Matching Scheme | **Features** | Recognition Method | **Dataset** | **Accuracy** |
|---|---|---|---|---|---|---|
| [78] | Digit Recognition | Word | MFCC | Neural Network | Self-made | 92% |
| [73] | Speech to Text | Word | MFCC | DNN | [49] | 99.08% |
| [80] | Digit Recognition | Word | MFCC | Sphinx-4 [84] | [42] | 85% |
| [85] | Speech to Text | Phoneme | MFCC | DNN & HMM | Self-made | 54.7% |
| [79] | Digit Recognition | Word | MFCC | Deep Belief Network [86] | Self-made | 94% |
| [75] | Speech Command Recognition | Word | MFCC | CNN | Self-made | 74% |
| [81] | Digit Recognition | Word | MFCC | LSTM | [42] | 86.8% |
| [82] | Digit Recognition | Word | MFCC | CNN | Self-made | 98% |
| [69] | Speech to Phoneme | Phoneme | MFCC | AutoEncoder | Self-made | 82.5% |
| [87] | Speech to Text | Word | MFCC & Local Features | HMM | Self-made | 93.7% |
| [77] | Speech to Text | Word | Spectral Analysis | Feed-Forward Network | Self-made | 60% |
| [68] | Speech to Text | Word | MFCC & LPC | HMM & DTW | Self-made | 84% |
| [67] | Speech to Text | Phoneme | MFCC | HMM | Self-made | 88.6% |
| [76] | Speech to Text | Word | Raw wave | End-to-End Recurrent Network | [50] [44] | 59.7% |
| [64] | Speech to Text | Phoneme | MFCC | Sphinx-3 [88] | Self-made | 87% |
| [14] | Speech to Text | Phoneme | MFCC, LDA [89], & MLLT [90] | Kaldi [91] | Self-made | 94.6% |
| [72] | Speech to Text | Phoneme | MFCC, LDA, MLLT | GMM, DNN, HMM | Self-made | 96.04% |

A renowned Bengali search engine Pipilika [71] developed a Bengali ASR system that used a larger vocabulary and performed better than previous DNN based methods [72]. Hybrid models combining DNN-HMM and GMM-HMM were also introduced and proved to perform better than previously applied architectures [73]. The DNN-GMM model firstly performed GMM, and the outputs of the GMM states were passed to DNN fully connected layers. An error pattern analysis of HMM is also analyzed for Bengali speech [74]. Similar efforts are given in speech to word ASR [68, 75, 76], phoneme-based ASR [77], spoken digit recognition systems [78, 79, 80, 81, 82], and word segmentation system [83].

Table 3 gives a detailed insight into the various architectures that are implemented in the scope of the Bengali Speech Recognition system. Although the paper focuses on the speech-to-text procedures, all methods that only operate in speech recognition (does not perform text translation) are evaluated. From the presented data, it can be observed that most methods are implemented using self-made datasets, which in most cases are inadequate in size. Therefore, the results presented in most works remain incompetent in performance standards. Further, Figure 6 illustrates a taxonomy of the implemented system toward Bengali ASR. Among the various methods, most of them implement phoneme-level recognition, which is the smallest and simplistic possible speech recognition level.

In Bengali ASR, no research works have performed an in-depth analysis of language-dependent challenges of ASR systems. Therefore, research works introduce their strength by showing better Word Error Rate (WER) and solving language-independent tasks. Table 4 shows the generated report based on our analysis of solving language-independent challenges. However, the investigation lacks all papers described in Table 3 as they do not disclose any parameters properly in the reports. The investigation shows that [69] and [76] have solved all the language-independent challenges. However, [69] only performed phoneme classification, and [76] concluded that the implemented architecture gave unsatisfactory WER. Hence, this study proves that the Bengali ASR systems have not converged to the acceptance level.

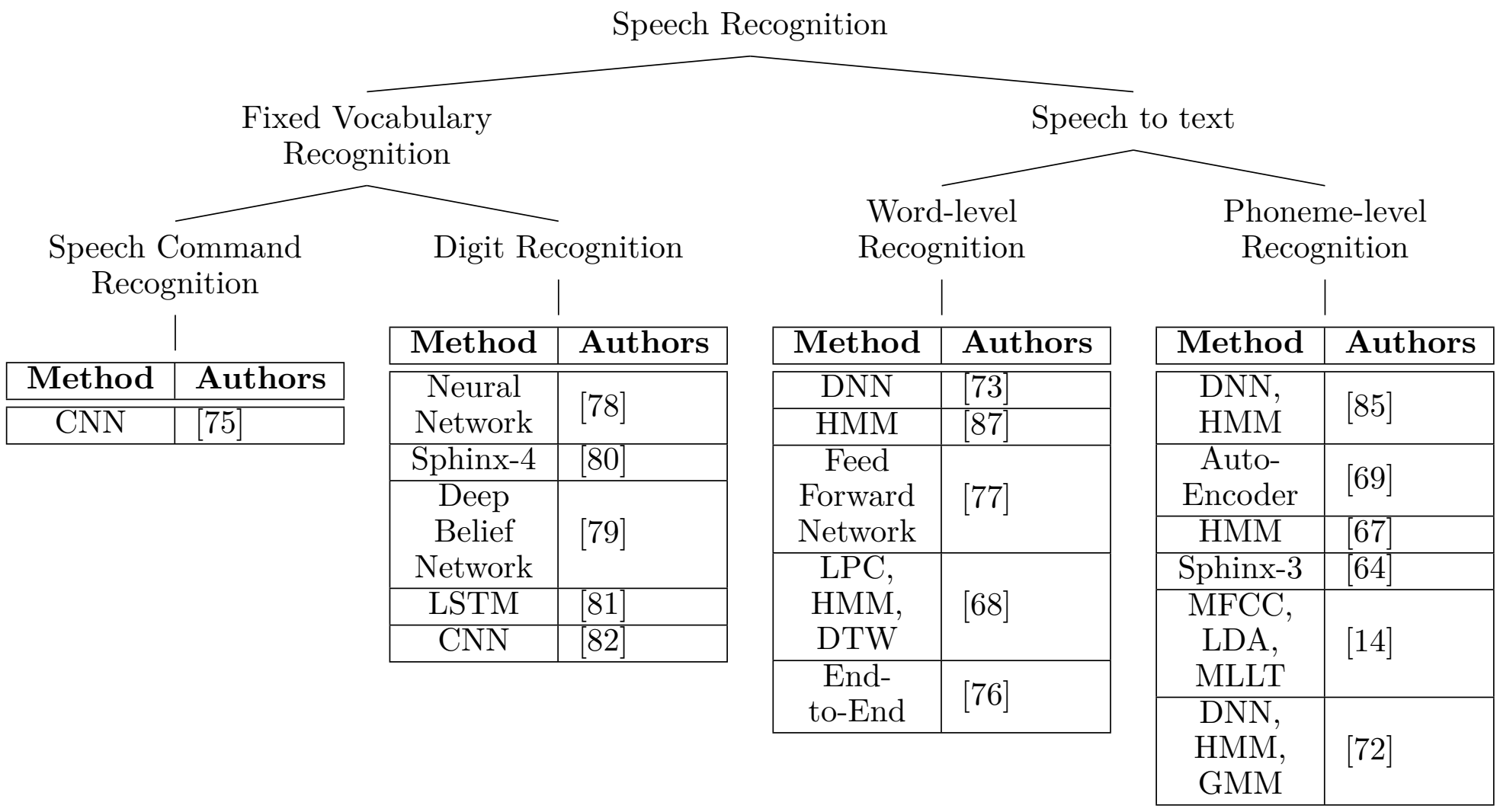

Figure 6: The figure illustrates a taxonomy of the papers in the domain of Bengali ASR. The taxonomy separates existing strategies firstly, based on the vocabulary limit and secondly, based on recognition level.

Table 4: The analysis report is conducted based on the language-independent challenges. The tick mark (✓) ensures the specific challenge is solved, the cross mark (✗) defines that the specific challenge is unsolved, and the unreported information is marked as null.

| **Method** | **Noise Reduction** | **Speaker Independent** | **Speech Variability** | **Speech Segmentation** | **Recording** |
|---|---|---|---|---|---|
| [79] | ✗ | ✓ | ✗ | ✗ | 8192Hz |
| [80] | ✗ | ✓ | ✗ | ✓ | null |
| [69] | ✓ | ✓ | ✓ | ✓ | 16000Hz |
| [81] | ✓ | ✓ | ✗ | ✓ | null |
| [72] | ✗ | ✓ | ✗ | ✓ | null |
| [76] | ✓ | ✓ | ✓ | ✓ | 16000Hz, single channel |
| [75] | ✗ | ✓ | ✗ | ✗ | null |
| [77] | ✗ | ✓ | ✗ | ✓ | null |
| [73] | ✗ | ✓ | ✓ | ✓ | 16000Hz, 16 bit, mono channel |
| [82] | ✗ | ✓ | ✓ | ✗ | null |

In the next section, the explicit challenges that the Bengali ASR system casts are addressed. To investigate the challenges of the Bengali ASR system, we assume the system contains the following properties:

- The end-to-end system must take speech as input, and the output must be in Bengali text.
- The Bengali ASR system will process continuous speech. The system will continuously get user voice input and segment the speech from the voice input and perform recognition.
- Speech may represent meaningful or meaningless words as human names are often out of the scope of Bengali vocabulary. However, word-matching ASR systems would fail to recognize meaningless words.

## 4 Challenges of Speech Recognition for Bengali

The difficulties of speech recognition can be split into two sections, a) language-dependent challenges and b) language-independent challenges. The principal processing architectures must be designed considering these challenges, and resolving these issues will cause a Bengali ASR system to perform better. The language-independent challenges are, a) noise, b) speaker dependency, c) speech variability, d) speech segmentation, and e) recording device. Contrastively, the language-dependent challenges are, a) structural properties, b) consonant conjuncts, c) diacritics, d) word database, e) dialects, f) silent letters, g) word utterance similarity, h) letter utterance similarity. Figure 7 illustrates the dependency of the hurdles.

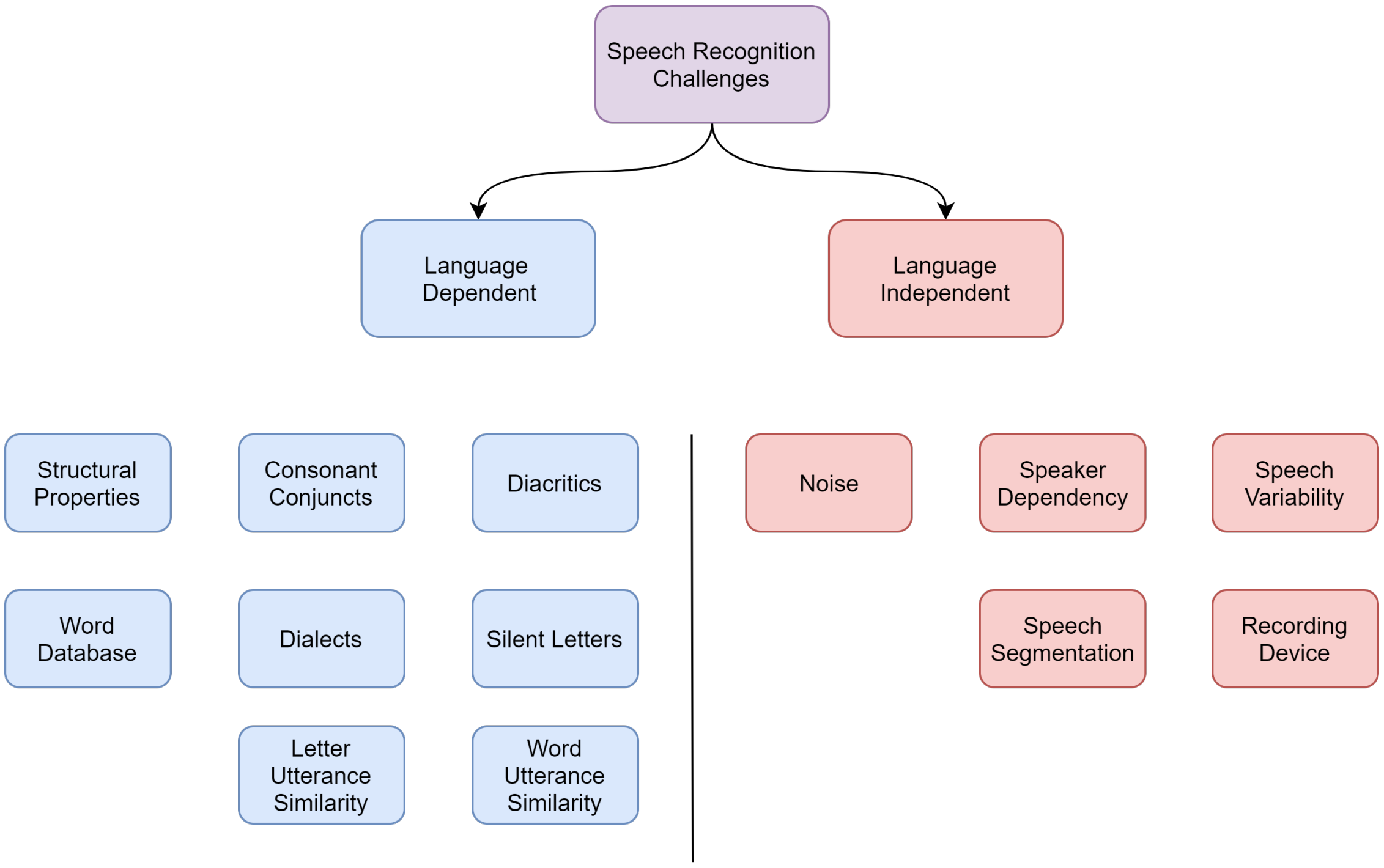

Figure 7: The overall challenges of speech recognition systems.

The researchers adequately perceive the language-independent speech-recognition challenges, and there exist state-of-the-art methods to suppress the difficulties. Also, efforts have been made to demonstrate the language-independent difficulties of speech recognition techniques [6, 92] and feature extraction procedures [93, 94]. Herefore, in the following subsections, we manifest the language-dependent difficulties that are overlooked concerning a Bengali speech recognition system and report some possible solutions. Nevertheless, we shortly define the language-independent challenges in Table 5.

### 4.1 Structural Properties

Every language has its structural properties that differ from language to language. Structural properties define the construction criteria of a meaningful sentence, which is set by grammar. However, languages composed of the same states or continentals hold similar grammatical structures, linguistic patterns, and writing patterns. In this case, the Bengali language has a significant share of relation in grammatical structure to the Hindi language. To reveal the structural properties of Bengali language, some of the fundamental structural differences between English and Bengali sentence are reported as follows,

- **Difference in Sentence Pattern:** English sentences have a sentence structure as, subject + verb + object, whereas, Bengali sentences have, subject + object + verb.
- **Absence of Auxiliary Verbs:** The Bengali language does not have the usage of auxiliary verbs in sentences.

Table 5: A summary of the language-independent challenges.

| Challenge | Description |
|---|---|
| Noise | The environmental sound mixed with speech. Noise distorts the speech features and may cause incorrect word outputs. Therefore, noise reduction/elimination is an important factor in preprocessing. |
| Speaker Dependency | Speaker dependency targets speakers for an ASR system. If an ASR system is designed for a particular individual, it is considered as a speaker-dependent ASR system, otherwise a speaker-independent ASR system. Modern ASR systems are speaker-independent, and therefore, they are trained with speeches of different individuals. |
| Speech Variability | Speech variability describes the change of utterance depending on human emotion, environmental, and age factors. Proper ASR architectures trained with variable speech datasets can overcome this challenge. |
| Recording Device | The recording device fixes the audio type used for the ASR system. The input audio can be a single channel (mono), dual-channel, stereo, or even intensity stereo. Every input type has its advantages and disadvantages depending on circumstances which is also a challenge. |
| Speech Segmentation | Speech segmentation can be classified into two types: a) word segmentation, b) phoneme segmentation. Word and phoneme segmentation are required for continuous speech recognition. Error in segmentation causes misleads in speech pattern matching. However, some present end-to-end ASR systems do not require speech segmentation [95]. |

- **Preposition Placement:** A preposition is placed before a noun or a noun-equivalent word in the English language. However, In Bengali, the preposition is placed before the noun or noun-equivalent word if required.

Consonant = Consonant Base + Vowel

ক = ক্ + অ

Consonant Conjunct = Consonant Base + Consoant Base + Vowel

ক্ক = ক্ + ক্ + অ

Figure 8: The structural difference between consonants and consonant conjuncts along with examples.

Figure 9 shows a translation of an English sentence to Bengali. The translation has all the three properties mentioned above. The sentence pattern of the translated Bengali sentence is the same as mentioned above. It is to be noted that the auxiliary verb 'am' is missing in the Bengali sentence. Also, due to the prepositions 'of' and 'in' the noun equivalent words দরজার = দরজা + র and সামনে = সামনে + এ are merged with individual Vibhaktis (বিভক্তি) [96]. Due to these types of structural dissimilarities of Bengali and English, the architectures that perform excellently in English may not perform better for Bengali speech. Also, models like n-gram and recurrent networks may require the attention of the language structure. The requirement of n-gram or recurrent networks are presented in subsection 4.7.

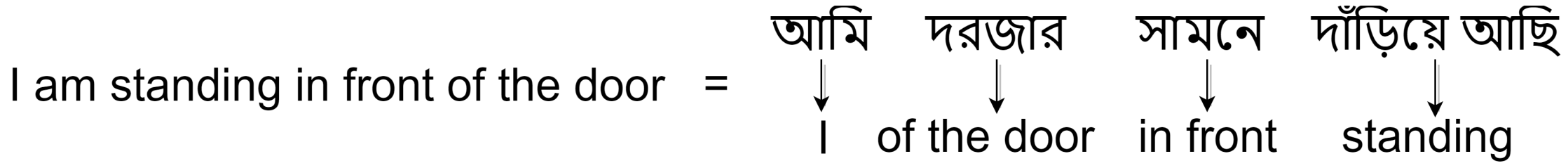

Figure 9: The structural difference between English and Bengali language.

### 4.2 Consonant Conjuncts

Consonant conjuncts are characters that hold two or more joined consonants represented as a single character. In the Bengali language, 118 consonant conjuncts are mostly used. Consonant conjuncts have been derived from the ancient Brahmi script, and it is also being used in many other scripts [97]. Figure 8 derives the difference between the utterance of consonant and consonant conjuncts. The utterance of consonants contains two portions, a consonant utterance followed by a vowel utterance. On the contrary, a consonant conjunct contains three portions, a consonant utterance followed by another consonant utterance, and finally, a vowel utterance. Consonant conjuncts may cause great difficulty to the phoneme-based speech recognition architecture. The precision of recognizing correct phoneme patterns must be ensured to recognize consonant conjuncts from the extracted speech features correctly.

### 4.3 Diacritics

In English, diacritics are practiced to express the correct accent of a word. Whereas, in the Bengali language, diacritics are greatly utilized to express words. The Bengali letters may contain at most two types of diacritics, vowel diacritics and consonant diacritics. The main difference between diacritics and consonant conjuncts is that diacritics are mostly considered an extension of a particular letter. On the contrary, consonant conjuncts are often considered as a single letter, and they can contain a diacritic as well. The diacritics are limited to 11 vowels and 7 consonants. In contrast, consonant conjuncts can be constructed with any pair of consonants. However, the second consonant can not be used as diacritics. The usage of diacritics introduces obstructions in phoneme matching and phoneme transformation processes.

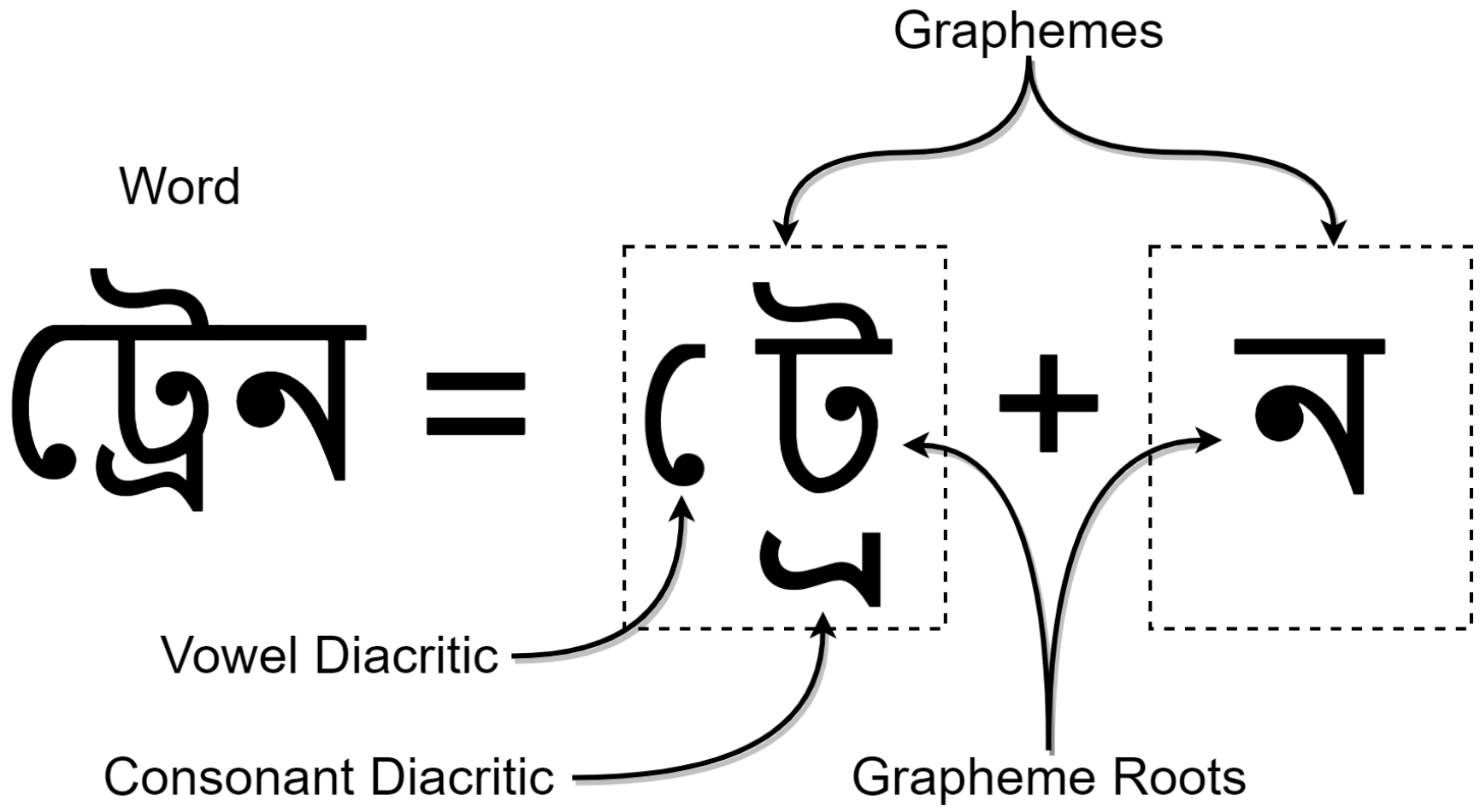

Figure 10: An illustration of word construction in Bengali language.

### 4.4 Word Database

A rich word database is one of the vastest language-dependent challenges of an ASR system. Word database is mainly required for an ASR system that uses speech-to-word identification. However, phoneme-based ASR systems are also trained using a word database, but they mostly learn to classify phonemes. The Bengali language has a complex structure of words due to the diacritics and consonant conjuncts. Figure 10 illustrates an example of the construction of Bengali words. Also, Figure 11 explains the construction of a grapheme, which is considered to be the smallest unit in a word writing system. A grapheme root can be obtained by excluding the diacritics from a grapheme. Diacritics and consonant conjuncts are the most critical challenge for a phoneme-based ASR system. Herefore, speech-to-word identification systems may be considered as a better choice. Nevertheless, due to the centuries of contact with the Europeans, Persians, Arabians, and Mughals, the Bengali vocabulary has a larger subset of adopted words. A linguistic difference is also considered in the Bengali and west Bengali continent. Therefore, generating a reliable speech-to-word database is also a significant challenge. A more extensive word database increases the probability of pattern mismatch. An incomplete word database will cause database-excluded words to be faultily recognized, mostly in word pattern matching ASR systems.

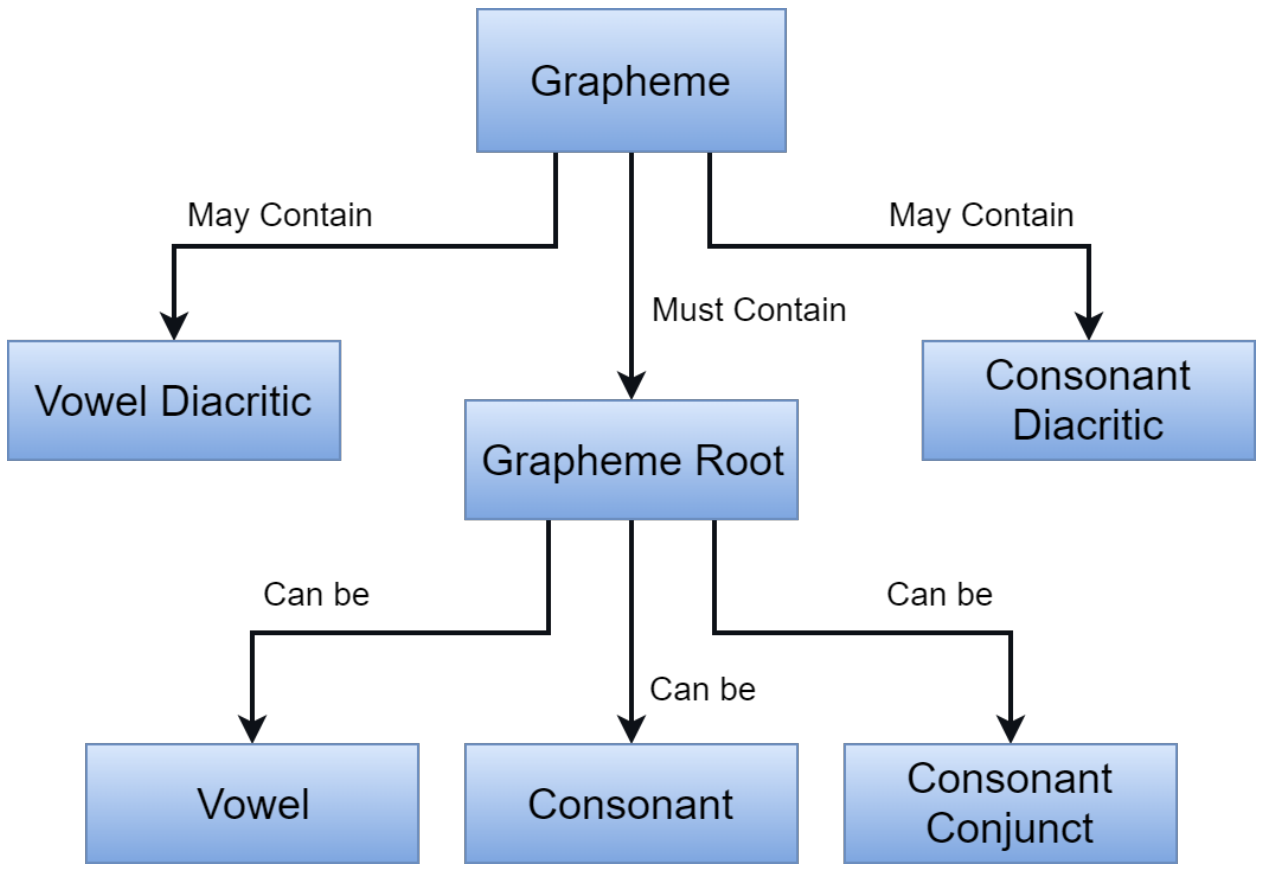

Figure 11: An illustration of all possible grapheme construction schemes in the Bengali language.

### 4.5 Dialects

Dialects refer to the linguistic variances that may differ in accent, vocabulary, spelling, and grammar of a language. Dialects are observed in almost every widely spoken language of the globe. According to the phonology and pronunciation of different dialects, the dialects of the Bengali language can be divided into six classes [98], a) Bengali, b) Rarhi, c) Varendri, d) Manbhumi, e) Rangpuri, and f) Sundarbani. Also, Bengali has more than 33 regional dialects. The dialects introduce more phoneme patterns and more words in vocabularies in an ASR system. These dialects should also be considered to implement a flawless Bengali ASR system.

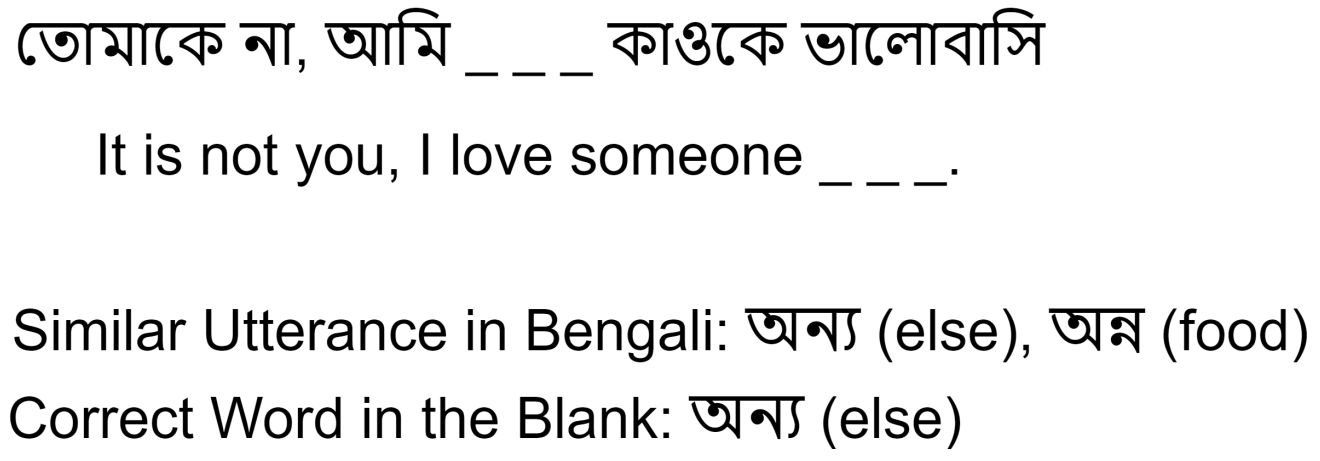

Figure 12: An example of guessing the correct word from a set of similar utterance words.

### 4.6 Silent Letters

Silent letters are frequently observed in most languages. Usually, a word containing letters that are not uttered is referred to as silent letters. Example: pneumonia (p silent) and ghost (h silent). Silent letters also occur in the Bengali language such as দুঃখ (ঃ silent), চাঁদ (ঁ silent). Speech to phoneme matching ASR system fails to recognize the silent letters. In this case, using a pre-defined lexicon rule can be used to auto-correct the words containing silent letters.

### 4.7 Word Utterance Similarity

In the Bengali language, some words have similar utterances but have different grapheme construction. As an illustration, the word pair অন্য (else), and অন্ন (food) has similar utterance, although their meaning is different. In such circumstances, humans mostly relate the correct word by the concept of the sentence and some intuition. This problem can essentially be solved using n-gram or recurrent neural network models over previous predictions. Figure 12 contains an illustration explaining the above scenario. From the example, from a set of similar uttered words, we humans pick the correct word by relating each word with the sentence. This explains the requirement of n-grams and recurrent networks over previously predicted words.

Table 6: The table illustrates phones of consonants in the Bengali language, along with letters, word examples (written in English), and the corresponding meanings. In the example, the words {জ্, য্}, {ন্, ণ্}, and {শ্, স্, ষ্} contain a similar phone structure.

| Phoneme | Letters | Bengali Word | Meaning |
|---|---|---|---|
| \k\ | ক্ | **k**apor | cloth |
| \k$^{h}$\ | খ্ | **kh**abar | food |
| \g\ | গ্ | **g**olap | rose |
| \g$^{h}$\ | ঘ্ | **gh**or | home |
| \ŋ\ | ঙ্ | ba**ng** | frog |
| \c\ | চ্ | **c**haka | tire |
| \c$^{h}$\ | ছ্ | **ch**ar | offer |
| \ɟ\ | জ্, য্ | **j**al | mesh |
| \ɟ$^{h}$\ | ঝ্ | **jh**al | hot taste |
| \t\ | ট্ | **t**aka | money |
| \t$^{h}$\ | ঠ্ | **th**ela | push |
| \d\ | ড্ | **d**al | branch |
| \d$^{h}$\ | ঢ্ | **dh**aka | covered |
| \t̪\ | ত্ | **t**ala | lock |
| \t̪$^{h}$\ | থ্ | **th**ana | police station |
| \d̪\ | দ্ | **d**alan | building |
| \d̪$^{h}$ \ | ধ্ | **dh**akka | push |
| \n\ | ন্, ণ্ | **n**am | name |
| \p\ | প্ | **p**oka | insect |
| \p$^{h}$\ | ফ্ | **ph**ol | fruit |
| \b\ | ব্ | **b**oka | fool |
| \b$^{h}$\ | ভ্ | **bh**ara | fare |
| \m\ | ম্ | **m**ash | month |
| \ʃ\ | শ্, স্, ষ্ | **sh**aban | soap |
| \r\ | র্ | **r**od | sun ray |
| \l\ | ল্ | **l**athi | stick |
| \h\ | হ্ | **h**ashi | smile |

### 4.8 Letter Utterance Similarity

Some Bengali letters also contain mostly similar utterances. Table 7 and 6 contain a list of vowels and consonants, their phone, Bengali word examples (written in English), and the meanings, respectively. Based on the example of Table 7 and 6, it can be observed that Bengali language contains some phonetically similar word clusters ({উ, ঊ}, {ই, ঈ}, {জ্, য্}, {ন্, ণ্}, and {শ্, স্, ষ্}). Also, humans often tend to fail to guess the correct letters from these clusters applied in particular words. For example, a vowel utterance "u" can be constructed using two different letters উ, ঊ. However, for a particular Bengali word, "ch**u**l" the correct word construction is "চ্ + উ + ল " (implicated in Table 7, row 4). Further similar variation is observed for consonant letters as well. An example can be drawn for the letter cluster ন্, ণ্. The consonant utterance "n" can be constructed using either ন্ or , ণ্. However, a specific word "**n**am" has a fixed word construction "ন্ + আ + ম" (shown in Table 6, row 18). The problem can be resolved either by applying a robust dataset that can give the pattern matcher a proper intuition or hard-implementing the Bengali grammatical rules (ণত্ব - ষত্ব বিধান, Bangla Academy laws) [99, 100].

Table 7: The table illustrates phones of vowels in the Bengali language, along with letters, word examples (written in English), and the corresponding meanings. In the example, the words {উ, ঊ} and {ই, ঈ} contain a similar phone structure.

| Phoneme | Letters | Bengali Word | Meaning |
|---|---|---|---|
| \ɑ\ | অ | **o**lpo | less |
| \ã\ | আ | **aa**mar | my |
| \i\ | ই, ঈ | **i**t**i**hash | history |
| \u\ | উ, ঊ | ch**u**l | hair |
| \e\ | এ | k**e** | who |
| \o\ | ও | g**o**lap | rose |
| \ou\ | ঔ | k**ou**shol | strategy |

## 5 Future Research Scope on Bengali ASR

This section summarizes the key challenges of a Bengali ASR system required to further extend the existing schemes' performance. Moreover, we propose an architecture that may solve the challenges. From the overall discussion of Section 4, three essential language-dependent challenges can be summarized:

- **Grammatical and literal dependency of words:** The grammatical dependency of words causes filter-out words depending on the grammatical structure of the previous words. Furthermore, literal dependency may help to obtain the proper word from a set of similar words containing a similar utterance pattern. Therefore, the search space for the proper word can be reduced. However, a powerful memory-based architecture is required to extract grammatical and literal dependencies properly. An attempt to implement this scheme may result in solving the challenges discussed in Section 4.1, 4.4, 4.5, and 4.7.
- **Grammatical and preceding dependency of characters:** The grammatical and preceding dependency deals with exploring the correct graphemes, vowel diacritics, and consonant diacritics of a word. Every language has grammatical patterns that correctly guess the proper graphemes from a grapheme set of a similar utterance. The extraction of these patterns also requires a memory-based generator. An attempt to implement this scheme may result in solving the challenges discussed in Section 4.2, 4.3, and 4.8.
- **Dissimilar uttered words due to silent letters:** Through the discussion in Section 4.6, it can be observed that dissimilarity of utterance and text mainly occurs due to silent letters. In the scope of the Bengali language, silent letters mostly do not contain grammatical dependencies. Therefore, a direct word-to-text transition may result in solving the difficulty.

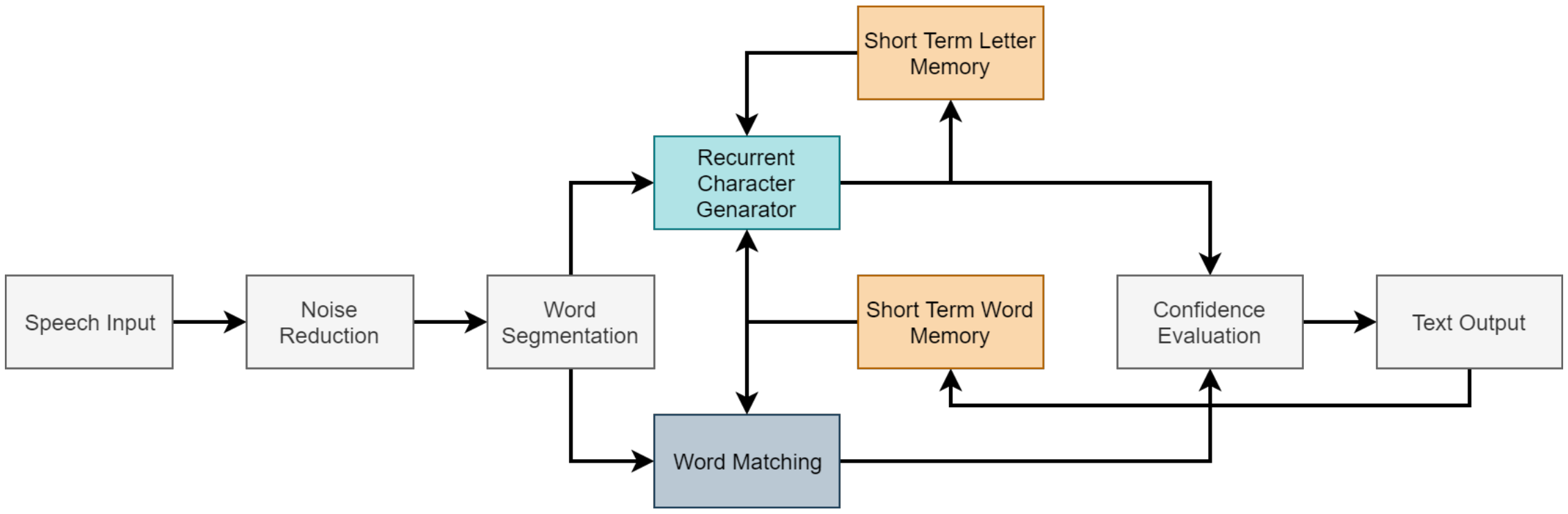

Figure 13: The suggested architecture of an optimal Bengali ASR system. Combining short-term memories will enable the architecture to recognize both words and characters' grammatical and literary dependency. Word matching schemes may help to recognize words that contain dissimilar and silent letters. Confidence evaluation defines if the model is confident that a speech exits in the present speech to word dictionary. Otherwise, the model extracts recurrent characters based on the speech.

The present researches in Bengali ASR systems often evade the interrelation of the grammatical issues and correct word predictions. Therefore the problems mentioned above are the future research scope of the Bengali ASR. Furthermore, we contribute to the future scope of the Bengali ASR by proposing a theoretical architecture. In Figure 13, we introduce an architecture that we believe to be optimal based on our research endeavor. To the best of our knowledge, the suggested architecture has not been investigated or implemented in any research endeavors. Also, the proposed architecture includes recurrent hybrid architecture that can create a new architectural perspective in the current research field. Hence, we point out the properties of the suggested ASR system as follows.

- The grammatical dependency of words mostly serves to find optimal literary words by generating some fixed rules. Short-term memory can be combined to correlate these rules. Using the short term memory, the system can optimally learn the grammatical relation only if trained on a large speech corpus.
- The grammatical and preceding dependency of characters can also be determined by combining a short-term memory with a speech character generator. The popular systems [36] depend on short-term memory to explore the dependency of character-level prediction.
- Every language, including the Bengali, contains words with irregular letter sequences. This problem can be solved by memorizing some fixed words. Therefore, it is optimal to implement both speech-to-word matching and phoneme-to-word matching. The current architectures implement end-to-end schemes [101, 102] that only generates characters and receives information from the previous characters only. Therefore it is usual to overlook most of the irregular word representations.
- The current implementations [36] only emphasize character recognition schemes. However, a hybrid implementation of a word and character matching scheme can solve the problem of generating irregular words and non-dictionary words. Therefore, our suggested system may search for optimal word matching. Further, the model may extract characters from the speech if the optimal word match is not found.

The suggested architecture pattern may solve the overall challenges discussed in the paper only if it is trained with speech corpora with a proper variation of speech and grammar variability.

## 6 Conclusion

In this survey, we begin with the investigation of the current research endeavors conducted in the Bengali ASR system, including speech corpora and recognition methods. Then, we have examined several difficulties that prevail in the domain of the Bengali ASR system. We have explained the structural and linguistic dissimilarities between languages on which an ASR system researcher should concentrate. We have rigorously presented grammatical fundamentals and suggestions on solving challenges. Although the examined challenges are also witnessed for most other languages, we have explained the challenges and opportunities regarding the Bengali language in particular. We have also investigated most of the latest works that implemented Bengali ASR systems, and through onerous exploration, we have shown that they lack perfection. We strongly believe that our genteel excavation on this very topic may expand the research scope of the Bengali as well as universal ASR systems and guide researchers scrupulously to target the exact challenges to be resolved.

## References

[1] Lawrence R Rabiner. Selected applications in speech recognition. *Readings in speech recognition*, page 267, 1990.

[2] Edmondo Trentin and Marco Gori. A survey of hybrid ann/hmm models for automatic speech recognition. *Neurocomputing*, 37(1-4):91–126, 2001.

[3] Mohamed Benzeghiba, Renato De Mori, Olivier Deroo, Stephane Dupont, Teodora Erbes, Denis Jouvet, Luciano Fissore, Pietro Laface, Alfred Mertins, Christophe Ris, et al. Automatic speech recognition and speech variability: A review. *Speech communication*, 49(10-11):763–786, 2007.

[4] Sven L Mattys, Matthew H Davis, Ann R Bradlow, and Sophie K Scott. Speech recognition in adverse conditions: A review. *Language and Cognitive Processes*, 27(7-8):953–978, 2012.

[5] Laurent Besacier, Etienne Barnard, Alexey Karpov, and Tanja Schultz. Automatic speech recognition for under-resourced languages: A survey. *Speech Communication*, 56:85–100, 2014.

[6] Ayushi Y Vadwala, Krina A Suthar, Yesha A Karmakar, Nirali Pandya, and Bhanubhai Patel. Survey paper on different speech recognition algorithm: Challenges and techniques. *Int. J. Comput. Appl.*, 175(1):31–36, 2017.

[7] RECEP SİNAN ARSLAN and NECAATTİN BARIŞÇI. A detailed survey of turkish automatic speech recognition. *Turkish Journal of Electrical Engineering & Computer Sciences*, 28(6):3253–3269, 2020.

[8] D Raj Reddy. Speech recognition by machine: A review. *Proceedings of the IEEE*, 64(4):501–531, 1976.

[9] Joe Tebelskis. *Speech recognition using neural networks.* PhD thesis, Carnegie Mellon University, 1995.

[10] Mark Gales, Steve Young, et al. The application of hidden markov models in speech recognition. *Foundations and Trends® in Signal Processing*, 1(3):195–304, 2008.

[11] CR Rashmi. Review of algorithms and applications in speech recognition system. *Int. J. Comput. Sci. Inf. Technol*, 5(4):5258–5262, 2014.

[12] Md Amaan Haque, Abhishek Verma, John Sahaya Rani Alex, and Nithya Venkatesan. Experimental evaluation of cnn architecture for speech recognition. In *First International Conference on Sustainable Technologies for Computational Intelligence*, pages 507–514. Springer, 2020.

[13] Toktam Zoughi, Mohammad Mehdi Homayounpour, and Mahmood Deypir. Adaptive windows multiple deep residual networks for speech recognition. *Expert Systems with Applications*, 139:112840, 2020.

[14] Soma Khan, Madhab Pal, Joyanta Basu, Milton Samirakshma Bepari, and Rajib Roy. Assessing performance of bengali speech recognizers under real world conditions using gmm-hmm and dnn based methods. In *SLTU*, pages 192–196, 2018.

[15] Tetsuya Takiguchi and Yasuo Ariki. Pca-based speech enhancement for distorted speech recognition. *Journal of multimedia*, 2(5), 2007.

[16] Reinhold Haeb-Umbach and Hermann Ney. Linear discriminant analysis for improved large vocabulary continuous speech recognition. In *Proc. ICASSP*, volume 1, pages 13–16. USA: ICASSP, 1992.

[17] Oh-Wook Kwon and Te-Won Lee. Phoneme recognition using ica-based feature extraction and transformation. *Signal Processing*, 84(6):1005–1019, 2004.

[18] Mariusz Ziółko, Rafał Samborski, Jakub Gałka, and Bartosz Ziółko. Wavelet-fourier analysis for speaker recognition. In *17th National Conference on Applications of Mathematics in Biology and Medicine*, volume 134, page 129, 2011.

[19] Najim Dehak, Patrick J Kenny, Réda Dehak, Pierre Dumouchel, and Pierre Ouellet. Front-end factor analysis for speaker verification. *IEEE Transactions on Audio, Speech, and Language Processing*, 19(4):788–798, 2010.

[20] Ehsan Variani, Xin Lei, Erik McDermott, Ignacio Lopez Moreno, and Javier Gonzalez-Dominguez. Deep neural networks for small footprint text-dependent speaker verification. In *2014 IEEE International Conference on Acoustics, Speech and Signal Processing (ICASSP)*, pages 4052–4056. IEEE, 2014.

[21] Fang Zheng, Guoliang Zhang, and Zhanjiang Song. Comparison of different implementations of mfcc. *Journal of Computer science and Technology*, 16(6):582–589, 2001.

[22] Chadawan Ittichaichareon, Siwat Suksri, and Thaweesak Yingthawornsuk. Speech recognition using mfcc. In *International Conference on Computer Graphics, Simulation and Modeling (ICGSM'2012)*, pages 28–29, 2012.

[23] Martin Westphal. The use of cepstral means in conversational speech recognition. In *Fifth European Conference on Speech Communication and Technology*, 1997.

[24] Hynek Hermansky and Nelson Morgan. Rasta processing of speech. *IEEE transactions on speech and audio processing*, 2(4):578–589, 1994.

[25] Hynek Hermansky and Petr Fousek. Multi-resolution rasta filtering for tandem-based asr. Technical report, IDIAP, 2005.

[26] KI Molla and Keikichi Hirose. On the effectiveness of mfccs and their statistical distribution properties in speaker identification. In *2004 IEEE Symposium on Virtual Environments, Human-Computer Interfaces and Measurement Systems, 2004.(VCIMS).*, pages 136–141. IEEE, 2004.

[27] Namrata Dave. Feature extraction methods lpc, plp and mfcc in speech recognition. *International journal for advance research in engineering and technology*, 1(6):1–4, 2013.

[28] Santosh K Gaikwad, Bharti W Gawali, and Pravin Yannawar. A review on speech recognition technique. *International Journal of Computer Applications*, 10(3):16–24, 2010.

[29] Jordan J Bird, Elizabeth Wanner, Anikó Ekárt, and Diego R Faria. Phoneme aware speech recognition through evolutionary optimisation. In *Proceedings of the Genetic and Evolutionary Computation Conference Companion*, pages 362–363, 2019.

[30] Kartik Audhkhasi, Brian Kingsbury, Bhuvana Ramabhadran, George Saon, and Michael Picheny. Building competitive direct acoustics-to-word models for english conversational speech recognition. In *2018 IEEE International Conference on Acoustics, Speech and Signal Processing (ICASSP)*, pages 4759–4763. IEEE, 2018.

[31] Alex Graves, Navdeep Jaitly, and Abdel-rahman Mohamed. Hybrid speech recognition with deep bidirectional lstm. In *2013 IEEE workshop on automatic speech recognition and understanding*, pages 273–278. IEEE, 2013.

[32] Herve A Bourlard and Nelson Morgan. *Connectionist speech recognition: a hybrid approach*, volume 247. Springer Science & Business Media, 2012.

[33] Alex Graves and Navdeep Jaitly. Towards end-to-end speech recognition with recurrent neural networks. In *International conference on machine learning*, pages 1764–1772, 2014.

[34] Awni Hannun, Carl Case, Jared Casper, Bryan Catanzaro, Greg Diamos, Erich Elsen, Ryan Prenger, Sanjeev Satheesh, Shubho Sengupta, Adam Coates, et al. Deep speech: Scaling up end-to-end speech recognition. *arXiv preprint arXiv:1412.5567*, 2014.

[35] Jan K Chorowski, Dzmitry Bahdanau, Dmitriy Serdyuk, Kyunghyun Cho, and Yoshua Bengio. Attention-based models for speech recognition. In *Advances in neural information processing systems*, pages 577–585, 2015.

[36] Linhao Dong, Shuang Xu, and Bo Xu. Speech-transformer: a no-recurrence sequence-to-sequence model for speech recognition. In *2018 IEEE International Conference on Acoustics, Speech and Signal Processing (ICASSP)*, pages 5884–5888. IEEE, 2018.

[37] Sepp Hochreiter and Jürgen Schmidhuber. Long short-term memory. *Neural computation*, 9(8):1735–1780, 1997.

[38] Junyoung Chung, Caglar Gulcehre, KyungHyun Cho, and Yoshua Bengio. Empirical evaluation of gated recurrent neural networks on sequence modeling. *arXiv preprint arXiv:1412.3555*, 2014.

[39] Kazuki Irie, Zoltán Tüske, Tamer Alkhouli, Ralf Schlüter, and Hermann Ney. Lstm, gru, highway and a bit of attention: An empirical overview for language modeling in speech recognition. In *Interspeech*, pages 3519–3523, 2016.

[40] Nimisha Srivastava, Rudrabha Mukhopadhyay, KR Prajwal, and CV Jawahar. Indicspeech: Text-to-speech corpus for indian languages. In *Proceedings of The 12th Language Resources and Evaluation Conference*, pages 6417–6422, 2020.

[41] Nimisha Srivastava, Rudrabha Mukhopadhyay, KR Prajwal, and CV Jawahar. IndicSpeech: Text-to-Speech Corpus for Indian Languages, 2021.

[42] Md Mahadi Hasan Nahid, Md. Ashraful Islam, Bishwajit Purkaystha, and Md Saiful Islam. Comprehending real numbers: Development of bengali real number speech corpus, 2018.

[43] Nahid, Md Mahadi Hasan. Bengali speech recognition - bangla real number audio dataset, 2018.

[44] Firoj Alam, SM Habib, Dil Afroza Sultana, and Mumit Khan. Development of annotated bangla speech corpora. 2010.

[45] Firoj Alam. Development of annotated bangla speech corpora, 2018.

[46] Biswajit Das, Sandipan Mandal, and Pabitra Mitra. Bengali speech corpus for continuous auutomatic speech recognition system. In *2011 International conference on speech database and assessments (Oriental COCOSDA)*, pages 51–55. IEEE, 2011.

[47] Biswajit Das, Sandipan Mandal, and Pabitra Mitra. SHRUTI Bengali Continuous ASR Speech Corpus, 2021.

[48] Mohi Reza, Warida Rashid, and Moin Mostakim. Prodorshok i: A bengali isolated speech dataset for voice-based assistive technologies: A comparative analysis of the effects of data augmentation on hmm-gmm and dnn classifiers. In *2017 IEEE Region 10 Humanitarian Technology Conference (R10-HTC)*, pages 396–399. IEEE, 2017.

[49] Sandipan Mandal, Biswajit Das, Pabitra Mitra, and Anupam Basu. Developing bengali speech corpus for phone recognizer using optimum text selection technique. In *2011 International Conference on Asian Language Processing*, pages 268–271. IEEE, 2011.

[50] Mark JF Gales, Kate M Knill, Anton Ragni, and Shakti P Rath. Speech recognition and keyword spotting for low-resource languages: Babel project research at cued. In *Fourth International Workshop on Spoken Language Technologies for Under-Resourced Languages (SLTU-2014)*, pages 16–23. International Speech Communication Association (ISCA), 2014.

[51] Mark JF Gales, Kate M Knill, Anton Ragni, and Shakti P Rath. IARPA Babel Bengali Language Pack, 2021.

[52] Google. *Large Bengali ASR training data set.*

[53] Shafayat Ahmed, Nafis Sadeq, Sudipta Saha Shubha, Md Nahidul Islam, Muhammad Abdullah Adnan, and Mohammad Zuberul Islam. Preparation of bangla speech corpus from publicly available audio & text. In *Proceedings of The 12th Language Resources and Evaluation Conference*, pages 6586–6592, 2020.

[54] Abu Quwsar Ohi, MF Mridha, Md Abdul Hamid, and Muhammad Mostafa Monowar. Deep speaker recognition: Process, progress, and challenges. *IEEE Access*, 9:89619–89643, 2021.

[55] Vassil Panayotov, Guoguo Chen, Daniel Povey, and Sanjeev Khudanpur. Librispeech: an asr corpus based on public domain audio books. In *2015 IEEE international conference on acoustics, speech and signal processing (ICASSP)*, pages 5206–5210. IEEE, 2015.

[56] R Karim, Md Shahidur Rahman, and Md Zafar Iqbal. Recognition of spoken letters in bangla. In *Proc. 5th international conference on computer and information technology (ICCIT02)*, 2002.

[57] AKMM Houque. Bengali segmented speech recognition system. *Undergraduate thesis, BRAC University, Bangladesh*, 2006.

[58] Md Rabiul Islam, Abu Sayeed Md Sohail, Md Waselul Haque Sadid, and MA Mottalib. Bangla speech recognition using three layer back-propagation neural network. In *Proceedings of the National Conference on Computer Processing of Bangla (NCCPB), Dhaka*, 2005.

[59] Md Rafiul Hassan, Baikunth Nath, and Mohammed Alauddin Bhuiyan. Bengali phoneme recognition: a new approach. In *Proc. 6th international conference on computer and information technology (ICCIT03)*, 2003.

[60] KJ Rahman, MA Hossain, D Das, T Islam, and MG Ali. Continuous bangla speech recognition system. In *Proc. 6th international conference on computer and information technology (ICCIT03)*, pages 303–307, 2003.

[61] Md Farukuzzaman Khan and Dr Ramesh Chandra Debnath. Comparative study of feature extraction methods for bangla phoneme recognition. In *5th ICCIT*, pages 27–28, 2002.

[62] Anup Kumar Paul, Dipankar Das, and Md Mustafa Kamal. Bangla speech recognition system using lpc and ann. In *2009 Seventh International Conference on Advances in Pattern Recognition*, pages 171–174. IEEE, 2009.

[63] K-F Lee, H-W Hon, and Raj Reddy. An overview of the sphinx speech recognition system. *IEEE Transactions on Acoustics, Speech, and Signal Processing*, 38(1):35–45, 1990.

[64] Sandipan Mandal, Biswajit Das, and Pabitra Mitra. Shruti-ii: A vernacular speech recognition system in bengali and an application for visually impaired community. In *2010 IEEE Students Technology Symposium (TechSym)*, pages 229–233. IEEE, 2010.

[65] Md Mijanur Rahman, Md Farukuzzaman Khan, and Mohammad Ali Moni. Speech recognition front-end for segmenting and clustering continuous bangla speech. *Daffodil International University Journal of Science and Technology*, 5(1):67–72, 2010.

[66] Shaheena Sultana, MAH Akhand, Prodip Kumer Das, and MM Hafizur Rahman. Bangla speech-to-text conversion using sapi. In *2012 International Conference on Computer and Communication Engineering (ICCCE)*, pages 385–390. IEEE, 2012.

[67] Foyzul Hassan, Mohammad Saiful Alam Khan, Mohammed Rokibul Alam Kotwal, and Mohammad Nurul Huda. Gender independent bangla automatic speech recognition. In *2012 International Conference on Informatics, Electronics & Vision (ICIEV)*, pages 144–148. IEEE, 2012.

[68] Md Akkas Ali, Manwar Hossain, Mohammad Nuruzzaman Bhuiyan, et al. Automatic speech recognition technique for bangla words. *International Journal of Advanced Science and Technology*, 50, 2013.

[69] Tanmay Bhowmik, Amitava Choudhury, and Shyamal Kumar Das Mandal. Deep neural network based recognition and classification of bengali phonemes: A case study of bengali unconstrained speech. In *International Conference on Next Generation Computing Technologies*, pages 750–760. Springer, 2017.

[70] Tanmay Bhowmik, Amitava Chowdhury, and Shyamal Kumar Das Mandal. Deep neural network based place and manner of articulation detection and classification for bengali continuous speech. *Procedia Computer Science*, 125:895–901, 2018.

[71] Shahjalal University of Science & Technology SUST. *Pipilika: (Bengali Search Engine)*, Accessed April 1, 2020.

[72] Jillur Rahman Saurav, Shakhawat Amin, Shafkat Kibria, and M Shahidur Rahman. Bangla speech recognition for voice search. In *2018 International Conference on Bangla Speech and Language Processing (ICBSLP)*, pages 1–4. IEEE, 2018.

[73] Md Alif Al Amin, Md Towhidul Islam, Shafkat Kibria, and Mohammad Shahidur Rahman. Continuous bengali speech recognition based on deep neural network. In *2019 International Conference on Electrical, Computer and Communication Engineering (ECCE)*, pages 1–6. IEEE, 2019.

[74] Shourin R Aura, Md J Rahimi, and Oli L Baroi. Analysis of the error pattern of hmm based bangla asr. *International Journal of Image, Graphics and Signal Processing*, 12(1):1, 2020.

[75] Shakil Ahmed Sumon, Joydip Chowdhury, Sujit Debnath, Nabeel Mohammed, and Sifat Momen. Bangla short speech commands recognition using convolutional neural networks. In *2018 International Conference on Bangla Speech and Language Processing (ICBSLP)*, pages 1–6. IEEE, 2018.

[76] Sakhawat Hosain Sumit, Tareq Al Muntasir, MM Arefin Zaman, Rabindra Nath Nandi, and Tanvir Sourov. Noise robust end-to-end speech recognition for bangla language. In *2018 International Conference on Bangla Speech and Language Processing (ICBSLP)*, pages 1–5. IEEE, 2018.

[77] Md Shafiul Alam Chowdhury and Md Farukuzzaman Khan. Linear predictor coefficient, power spectral analysis and two-layer feed forward network for bangla speech recognition. In *2019 IEEE International Conference on System, Computation, Automation and Networking (ICSCAN)*, pages 1–6. IEEE, 2019.

[78] Md Hossain, Md Rahman, Uzzal Kumar Prodhan, Md Khan, et al. Implementation of back-propagation neural network for isolated bangla speech recognition. *arXiv preprint arXiv:1308.3785*, 2013.

[79] Mahtab Ahmed, Pintu Chandra Shill, Kaidul Islam, Md Abdus Salim Mollah, and MAH Akhand. Acoustic modeling using deep belief network for bangla speech recognition. In *2015 18th International Conference on Computer and Information Technology (ICCIT)*, pages 306–311. IEEE, 2015.

[80] Md Mahadi Hasan Nahid, Md Ashraful Islam, and Md Saiful Islam. A noble approach for recognizing bangla real number automatically using cmu sphinx4. In *2016 5th International Conference on Informatics, Electronics and Vision (ICIEV)*, pages 844–849. IEEE, 2016.

[81] Md Mahadi Hasan Nahid, Bishwajit Purkaystha, and Md Saiful Islam. Bengali speech recognition: A double layered lstm-rnn approach. In *2017 20th International Conference of Computer and Information Technology (ICCIT)*, pages 1–6. IEEE, 2017.

[82] Riffat Sharmin, Shantanu Kumar Rahut, and Mohammad Rezwanul Huq. Bengali spoken digit classification: A deep learning approach using convolutional neural network. *Procedia Computer Science*, 171:1381–1388, 2020.

[83] Tanmay Bhowmik and Shyamal Kumar Das Mandal. Prosodic word boundary detection from bengali continuous speech. *Language Resources and Evaluation*, pages 1–19, 2019.

[84] Willie Walker, Paul Lamere, Philip Kwok, Bhiksha Raj, Rita Singh, Evandro Gouvea, Peter Wolf, and Joe Woelfel. Sphinx-4: A flexible open source framework for speech recognition, 2004.

[85] Mohammed Rokibul Alam Kotwal, Manoj Banik, Qamrun Nahar Eity, Mohammad Nurul Huda, Ghulam Muhammad, and Yousef Ajami Alotaibi. Bangla phoneme recognition for asr using multilayer neural network. In *2010 13th International Conference on Computer and Information Technology (ICCIT)*, pages 103–107. IEEE, 2010.

[86] Geoffrey E Hinton. Deep belief networks. *Scholarpedia*, 4(5):5947, 2009.

[87] Sara Binte Zinnat, Razia Marzia Asheque Siddique, Md Imamul Hossain, Deen Md Abdullah, and Mohammad Nurul Huda. Automatic word recognition for bangla spoken language. In *2014 International Conference on Signal Propagation and Computer Technology (ICSPCT 2014)*, pages 470–475. IEEE, 2014.

[88] Paul Placeway, S Chen, Maxine Eskenazi, Uday Jain, Vipul Parikh, Bhiksha Raj, Mosur Ravishankar, Roni Rosenfeld, Kristie Seymore, M Siegler, et al. The 1996 hub-4 sphinx-3 system. In *Proc. DARPA Speech recognition workshop*, volume 97. Citeseer, 1997.

[89] Suresh Balakrishnama and Aravind Ganapathiraju. Linear discriminant analysis-a brief tutorial. In *Institute for Signal and information Processing*, volume 18, pages 1–8, 1998.

[90] Mark JF Gales. Maximum likelihood linear transformations for hmm-based speech recognition. *Computer speech & language*, 12(2):75–98, 1998.

[91] Daniel Povey, Arnab Ghoshal, Gilles Boulianne, Lukas Burget, Ondrej Glembek, Nagendra Goel, Mirko Hannemann, Petr Motlicek, Yanmin Qian, Petr Schwarz, et al. The kaldi speech recognition toolkit. In *IEEE 2011 workshop on automatic speech recognition and understanding*, number CONF. IEEE Signal Processing Society, 2011.

[92] Priyanka Sahu, Mohit Dua, and Ankit Kumar. Challenges and issues in adopting speech recognition. In *Speech and Language Processing for Human-Machine Communications*, pages 209–215. Springer, 2018.

[93] S Nivetha. A survey on speech feature extraction and classification techniques. In *2020 International Conference on Inventive Computation Technologies (ICICT)*, pages 48–53. IEEE, 2020.

[94] Amitoj Singh, Virender Kadyan, Munish Kumar, and Nancy Bassan. Asroil: a comprehensive survey for automatic speech recognition of indian languages. *Artificial Intelligence Review*, pages 1–32, 2019.

[95] Jui-Yang Hsu, Yuan-Jui Chen, and Hung-yi Lee. Meta learning for end-to-end low-resource speech recognition. In *ICASSP 2020-2020 IEEE International Conference on Acoustics, Speech and Signal Processing (ICASSP)*, pages 7844–7848. IEEE, 2020.

[96] U.K. Cakrabartī. *Bāṃlā bākyera padagucchera saṃgaṭhana.* Pramā Prakāśanī, 1992.

[97] Tuṅga and Sudhāṃśu Śekhara. *Bengali and Other Related Dialects of South Assam.* Mittal Publications, 1 edition, 1995.

[98] Bhasa Vidya Parishad. *Indian Journal of Linguistics.* Number v. 20. Bhasa Vidya Parishad., 2001.

[99] S.K. Chatterji. *Bhāshā-prakāśa Bāṅgālā byākaraṇa.* Rūpā, 1988.

[100] Bāṃlā Ekāḍemī (Bangladesh). *Bangla Academy Journal.* Number v. 21, no. 2 - v. 22, no. 2. Bangla Academy., 1995.

[101] Alexei Baevski, Yuhao Zhou, Abdelrahman Mohamed, and Michael Auli. wav2vec 2.0: A framework for self-supervised learning of speech representations. *Advances in Neural Information Processing Systems*, 33, 2020.

[102] Mirco Ravanelli, Philemon Brakel, Maurizio Omologo, and Yoshua Bengio. Light gated recurrent units for speech recognition. *IEEE Transactions on Emerging Topics in Computational Intelligence*, 2(2):92–102, 2018.